# The Interpretability Analysis of the Model Can Bring Improvements to the Text-to-SQL Task

Cong Zhang

China Life Insurance Company Limited, Jilin Province Branch

zhangcong@jl.e-chinalife.com

## Abstract

To elevate the foundational capabilities and generalization prowess of the text-to-SQL model in real-world applications, we integrate model interpretability analysis with execution-guided strategy for semantic parsing of WHERE clauses in SQL queries. Furthermore, we augment this approach with filtering adjustments, logical correlation refinements, and model fusion, culminating in the design of the CESQL model that facilitates conditional enhancement. Our model excels on the WikiSQL dataset, which is emblematic of single-table database query tasks, markedly boosting the accuracy of prediction outcomes. When predicting conditional values in WHERE clauses, we have not only minimized our dependence on data within the condition columns of tables but also circumvented the impact of manually labeled training data. Our hope is that this endeavor to enhance accuracy in processing basic database queries will offer fresh perspectives for research into handling complex queries and scenarios featuring irregular data in real-world database environments.

## 1 Introduction

Currently, AI technology is profoundly transforming the database landscape. Text-to-SQL, by innovating data provisioning to cater to the information retrieval and data analysis needs of a broader audience of everyday users, is emerging as a catalyst for propelling databases towards greater efficiency, collaboration, and intelligence.

In recent years, text-to-SQL solutions leveraging large autoregressive models have continually surpassed existing methods on benchmark datasets for multi-table complex queries (Zhu et al., 2024), such as Spider (Yu et al., 2018c) and BIRD (Li et al., 2023), attributed to their exceptional natural language understanding and generation capabilities.

In reality, it is highly prevalent for users of reporting systems to conduct simple queries, statistical analyses, and evaluations on consolidated single-report data derived from multi-table integration and field augmentation within databases. The single-table query dataset exemplified by WikiSQL (Zhong et al., 2017) aligns well with this application scenario. Despite its relatively straightforward syntax and lesser complexity when compared to datasets like Spider and BIRD (Deng et al., 2022), WikiSQL continues to serve as a pivotal benchmark for demonstrating the technical feasibility of converting natural language into simple SQL and validating the fundamental capabilities of models.

# 2 Related work

Multiple models have achieved landmark breakthroughs in tackling the WikiSQL dataset processing task. Notably, SeaD (Xu et al., 2023) employs formatted input and output, converting the text-to-SQL task into a Seq2Seq task, and after applying data augmentation techniques such as Erosion and Shuffle, it successfully outputs accurate SQL queries and corresponding questions. SDSQL (Hui et al., 2021) reframes the text-to-SQL task into one akin to dependency relation prediction. IESQL (Ma et al., 2020), on the other hand, transforms the text-to-SQL task into mention extraction and linking. HydraNet (Qin et al., 2020) decomposes the text-to-SQL task into common NLP subtasks, thereby leveraging pre-trained models like BERT (Devlin et al., 2019) to its fullest potential.

## 2.1 Execution-guided strategy

The execution-guided strategy (Wang et al., 2018) is a prevalent decoding approach in the text-to-SQL field (Liu et al., 2025), designed to boost the accuracy of SQL query generation from natural language. By capitalizing on the unique features of SQL, this strategy facilitates the early execution of some candidate SQL queries. Based on the execution results, the model discards those queries that fail to execute or yield empty outcomes, thereby effectively reducing the likelihood of generating syntactically or semantically flawed SQL statements. Notably, the execution-guided strategy is employed not only by models that excel on the relatively straightforward single-table dataset of WikiSQL but also by CHASE-SQL (Pourreza et al., 2024) and RSL-SQL (Cao et al., 2024), which have demonstrated outstanding performance on the multi-table, complex datasets of Spider and BIRD.

## 2.2 The interpretability of the model

The interpretability of models constitutes a pivotal research area in deep learning, transforming black-box models into transparent ones, shedding light on the decision-making rationale of intricate machine learning models, and ultimately fostering our scientific comprehension of artificial intelligence itself. Notably, LIME (Local Interpretable Model-agnostic Explanations) (Ribeiro et al., 2016) and SHAP (SHapley Additive exPlanations) (Lundberg et al., 2017) stand out as two extensively adopted techniques for model interpretability.

LIME is a locally interpretable model-agnostic algorithm characterized by its relatively low computational complexity and robust versatility, making it applicable to various tasks associated with natural language processing. SHAP, an interpretable algorithm grounded in the Shapley value from game theory, facilitates both local and global interpretations by quantifying feature contributions. However, it tends to have a higher computational complexity.

# 3 Methods

The unenforceability of the execution-guided strategy, also termed runtime errors, can be proactively mitigated through rule validation. The empty output generated by the execution-guided strategy is not devoid of meaning, particularly when the conditional operator in the WHERE clause involves the greater-than or less-than sign, and the conditional value is numeric. It is frequently observed that the condition query yields no results in real-world situations, which does not necessarily constitute an error.

To better align with real-world scenarios, stay faithful to users' original query intentions, strike a balance between the model's semantic parsing results' reliance on the problem description and the data in the table, and offer a more rational candidate prediction value reference ranking for iterative execution of the guiding strategy, we have augmented the application of the guiding strategy with interpretability analysis and processing.

The complexity of large models presents a formidable challenge in tracing and comprehending their decision-making pathways, particularly due to the uncertainty inherent in the outputs of autoregressive large models, which exacerbates the difficulty of interpretability. Given the current limitation of computational power, we have chosen to employ LIME on BERT and its enhanced variant, SpanBERT (Joshi et al., 2020), which leverage an autoencoder architecture and offer greater interpretability, to facilitate the models' handling of the WikiSQL dataset and execution of text-to-SQL tasks.

We introduce the LIME library to develop a local surrogate model, which serves to approximate the decision-making behavior of a black-box model for conditional column classification predictions in the vicinity of specific texts. The weights assigned to feature units by the surrogate model directly mirror the degree to which each feature influences the prediction outcome, with their numerical values serving as a basis for determining conditional values or delineating boundaries. This process can be formalized as:

$$\omega = \arg\min_{\omega \in \mathbb{R}^n} \sum_{z \in Z(x)} \pi_x(z)[p(z) - \omega^\top z']^2$$

In this context, $x$ represents the original input sentence, which is a sequence of tokens with a length of $n$. $Z(x)$ denotes a set of perturbed sentences generated from $x$ through random masking or deletion of tokens. $z$ stands for a specific perturbed sentence used as input for the black-box model. $z'$ is a binary vector of the same dimension as $x$, where $z'_i$ equals 1 if the $i$th token is retained in $z$ and 0 if it is masked or deleted. $p(z)$ signifies the positive class probability obtained after inputting the perturbed sentence $z$ into the black-box model. $\pi_x(z)$ represents the neighborhood weight, which quantifies the similarity between the perturbed sentence $z$ and the original sentence $x$. $\omega$ denotes the weight vector to be determined, with $\omega_i$ indicating the contribution of the $i$th token to the positive class probability.

## 3.1 Comprehensive design

Our CE-SQL model divides the text-to-SQL task for processing the WikiSQL dataset into two modules: SELECT clause prediction and WHERE clause prediction. SELECT clause prediction encompasses column selection prediction and aggregate function prediction, while WHERE clause prediction comprises various subtasks, including condition column prediction, condition operator prediction, condition value prediction, and the prediction of the number of conditions.

The SELECT clause prediction module initially conducts joint predictions on the selected columns and their respective aggregation functions. Subsequently, for a small subset of prediction results containing syntactic errors (such as applying the SUM operation to text-type column data), it utilizes prompt templates to elicit hidden information and employs a fully connected layer for scoring via SOFTMAX. These results are then compared against the joint predictions, allowing for the correction of erroneous selected columns or aggregation functions through logical self-correction.

The WHERE clause prediction module utilizes entity relationship extraction to forecast triples comprising [condition column, condition operator, condition value]. When integrating various triple extraction models, one path of the models is selected, with the SIGMOID function threshold serving as a regulatory mechanism to manage the output of the entire triple array. If we consider parsing the relational operators between conditions in the WHERE clause, average pooling can be employed to capture global semantic information for three-category prediction. The prediction outcomes of conditional relational operators undergo external refinement based on the [condition column, condition operator, condition value] array, which has undergone selective execution guidance, interpretability analysis, and grammatical logic filtering.

When dealing with WHERE clauses where the condition column of the predicted triples is of numeric type, particularly when the corresponding condition operator is the greater than or less than sign, we endeavor to refrain from using execution-guided strategies and instead rely directly on interpretable feature statistics for verification. In cases where the condition column of the predicted triples within a WHERE clause is of text type, if the initial attempt using an execution-guided strategy yields an empty result, we can subsequently employ interpretable feature statistics to ascertain whether the condition triples require modification and, if so, the direction in which to make adjustments. This serves as a more comprehensive and intuitive reference for the continued cyclic execution of the guidance strategy.

### 3.2 Black-box model

We selected BERT and SpanBERT as black-box models for experimentation and discovered that SpanBERT outperformed BERT. When employing BERT as a black-box model, we computed weight values through weighted summation of feature units to ascertain their contribution to the prediction outcomes. When utilizing SpanBERT as a black-box model, we activated the hidden_states output to harness its span understanding capability, thereby enhancing the representation of crucial spans within the text and enriching the final output with more span-level semantics. Consequently, the derived weight values directly mirror the contribution of contiguous feature units to the prediction results.

### 3.3 Examples of application

In the task of extracting triples from WHERE clauses, particularly when the condition column is of text type, a common issue arises with the boundary confirmation of corresponding condition values. Notably, unlike other SQL decomposition tasks, condition values frequently encounter challenges where the extracted range is either excessively broad or insufficient. To address this, we employ the interpretability analysis of LIME, coupled with an execution-guided strategy, to validate the model's predictions. Let's delve into an example from the WikiSql dataset that demonstrates how to tackle such issues.

For the question "What is the grid total for Ralf Schumacher racing over 53 laps?", our model extracted a set of conditions in the WHERE clause, resulting in ['driver', 'is/are', 'Ralf Schumacher racing']. However, this result is incorrect, as the information unit extracted for the condition value exceeds what is expected in the standard answer. The extracted condition column field labeled 'driver' has an attribute of 'text'. If we utilize interpretability analysis for verification, we would first compile a statistics list of feature units related to it: [('What', 0.0229), ('is', -0.0011), ('the', -0.0070), ('grid', 0.0058), ('total', 0.0266), ('for', -0.0354), ('ralf', 0.1526), ('schumacher', 0.1618), ('racing', -0.3376), ('over', 0.0142), ('53', -0.0205), ('laps', -0.0118)]. Subsequently, we would compile another statistics list of feature units with varying widths (spans) that are partially related to the extracted value: [('ralf schumacher', 0.2493), ('schumacher', 0.1618), ('ralf', 0.1526),('ralf schumacher racing', 0.0846), ('schumacher racing', -0.0179), ('racing', -0.3376)]. The contribution value will serve as the guiding principle for iteratively executing the strategy, minimizing the chance of overlooking the correct answer. The fundamental principle is illustrated in Figure 1.

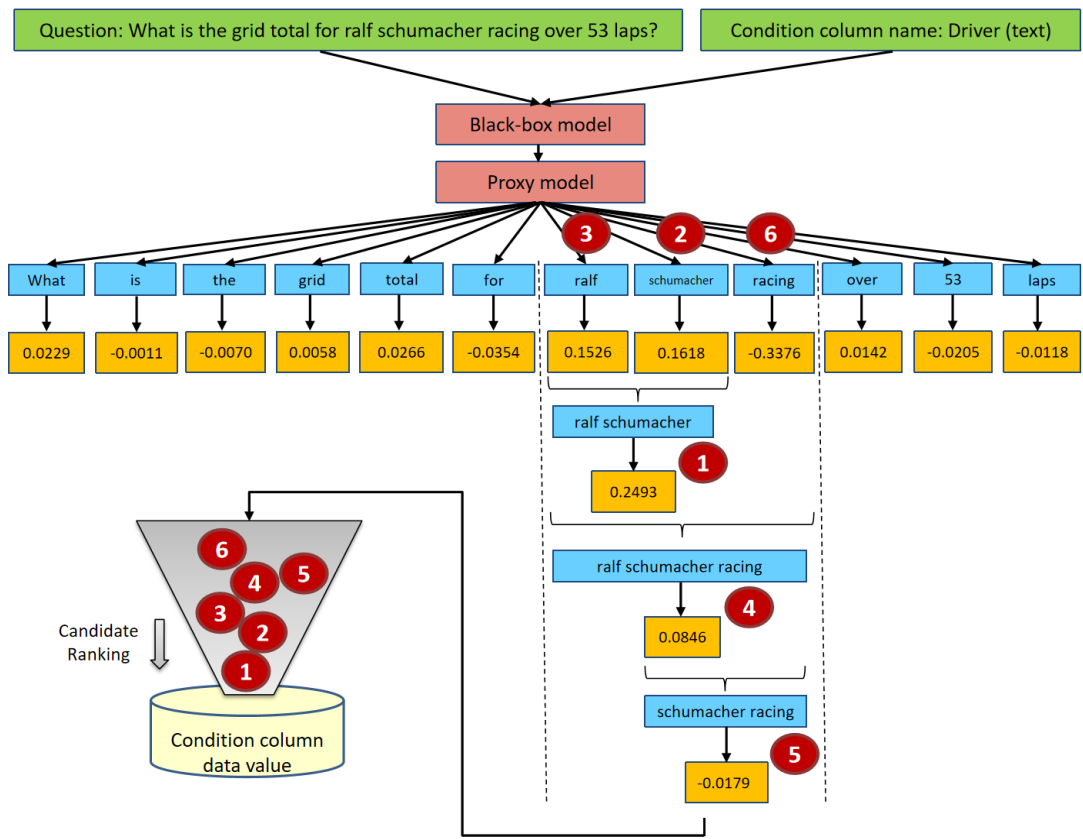

Figure 1: Analysis application of interpretability for WHERE clause condition values.

For the question "Name the casualties for the Kabul area?", we obtained a set of condition results from the WHERE clause as ['Location', 'is/are', 'Kabul'] through our model. The condition value does not align with the standard answer, suggesting inadequate information units have been extracted. The extracted condition column, labeled 'Location', has an attribute of 'text'. If we employ interpretability analysis for verification, we first compile a list of feature unit statistics related to it: [('Name', 0.0230), ('the', -0.0124), ('casualities', 0.1084), ('for', 0.1462), ('kabul', 0.3937), ('area', 0.0944]. Subsequently, we compute a list of feature unit statistics with varying widths (spans) that are partially associated with the extracted value: [('kabul area', 0.5722), ('kabul', 0.3937), ('area', 0.0944)]. We utilize the contribution value as the guiding principle for iteratively executing the strategy, thereby minimizing the likelihood of overlooking the correct answer.

In practical data queries, users often desire to swiftly ascertain the existence of data in a table that fulfills their specified conditions without having to peruse the entire database table. When the model analyzes such queries, particularly when dealing with conditions in a WHERE clause where the value of a numeric type condition column falls within a certain range, relying solely on an execution-guided strategy may lead to erroneous judgments. In such scenarios, the interpretability analysis offered by LIME can be instrumental in validating the model's prediction results. Let's examine an example addressing such issues within the WikiSql dataset.

For the question "What is the average ranking for a react with a value of 0.17300000000000001 and fewer than 5 lanes?", our model extracted a set of conditions in the WHERE clause, resulting in ['React', '=', '0.17300000000000001']. This aligns with the standard answer. However, since the number '0.17300000000000001' is absent from the data in the 'React' column, applying the execution-guided strategy for verification would lead to discarding this extraction result due to an empty query result based on this condition. The attribute of the 'React' field is 'real', indicating that the corresponding condition value should be numeric. If we utilize interpretability analysis for verification, we would compile a statistics list of feature units related to it: [('react', 0.5163), ('and', 0.1403), ('for', 0.0659), ('0.17300000000000001', 0.0625), ('ranking', 0.0441), ('a', 0.0435), ('What', 0.0247), ('than', 0.0099), ('the', 0.0017), ('less', -0.0007), ('5', -0.0074), ('of', -0.0079), ('average', -0.0116), ('is', -0.0179), ('lanes', -0.1020)]. By examining the statistical results for numerical information units, we found that the high-scoring number '0.17300000000000001' showed a positive correlation, while the low-scoring number '5' exhibited a negative correlation. Hence, the triple ['react', 'is/are', '0.17300000000000001'] is deemed correct.

For the question "Tell me the average spectators for June 21, 2006, with a time later than 21:00?", our model extracted a set of conditions from the WHERE clause, resulting in ['time(cet)', '>', '21']. This aligns with the standard answer. Nevertheless, since there is no data in the 'time(cet)' column exceeding 21, applying the execution-guided strategy for verification would lead to discarding this extraction due to an empty query result. The attribute of the 'time(cet)' field is 'real', indicating that the corresponding condition value should be numeric. If we utilize interpretability analysis for verification, we would compile a statistics list of feature units related to it: [('time', 0.4450), ('and', 0.1604), ('21', 0.1575), ('more', 0.0487), ('than', 0.0449), ('for', 0.0368), ('Tell', -0.0160), ('average', 0.0158), ('spectators', 0.0144), ('the', -0.0139), ('me', 0.0104), ('2006-06-21', -0.0072)]. Among the statistical results, we searched for numerical information units, finding that the number '21' scored high and was positively correlated, whereas '2006-06-21' scored low and was negatively correlated. Hence, we conclude that the triple ['time(cet)', '>', '21'] is accurate.

For the question "Who was the stage winner when the stage was smaller than 16, earlier than 1986, and a distance (km) was 19.6?", we extracted a set of condition results ['stage', 'less than', '1986'] from the WHERE clause using a model. The condition value does not align with the standard answer. However, considering that the data in the 'stage' condition column includes values less than 1986, if we employ the execution-guided strategy for verification, it will pass due to the existence of relevant data in the condition query. The attribute of the 'stage' field column is 'real', indicating that the corresponding condition value should be numeric. If we utilize interpretability analysis for verification, we would compile a statistics list of feature units related to it: [('stage', 0.4059), ('winner', 0.1751), ('than', 0.1079), ('16', 0.1036), ('when', 0.0563), ('and', 0.0368), ('smaller', 0.0251), ('earlier', 0.0085), ('the', 0.0041), ('a', 0.0024), ('1986', -0.0006), ('km', -0.0187), ('distance', -0.0213), ('Who', -0.0243), ('was', -0.0379), ('19.6', -0.0438)]. Among the statistical results, we search for numerical information units and find that the number '16' scores higher than '1986' and '19.6'. Therefore, we conclude that the triple ['stage', 'less than', '16'] is the correct one.

## 4 Experiments

In the course of our research, we employed the Bert4Keras, TensorFlow, and PyTorch frameworks. Given that we only had access to one GeForce RTX 3090 graphics card, we used the BERT-Base and SpanBERT-base models for some tasks, while opting for the BERT-Large model for others.

| Model | Dev Acc_lf | Dev Acc_ex | Test Acc_lf | Test Acc ex |
|---|---|---|---|---|
| HydraNet+EG | 86.6 | 92.4 | 86.5 | 92.2 |
| IESQL+EG ♣ | 85.8 | 91.6 | 85.6 | 91.2 |
| BRIDGE+EG ♦ | 86.8 | 92.6 | 86.3 | 91.9 |
| SDSQL+EG ♣ | 86.7 | 92.5 | 86.6 | 92.4 |
| SeaD+EG CS | 87.3 | 92.8 | 87.1 | 92.7 |
| CESQL+EG | **87.5** | **93.2** | **87.7** | **93** |

Table 1: Accuracy (%) of logic form Acc_lf and execution Acc_ex of our model CESQL and other competitors. Best results in bold. EG: execution-guided decoding. ♣ denotes methods that leverage additional annotation of dataset. ♦ denotes methods that utilize database content during training.

According to the comparative results presented in Table 1, it is evident that upon integrating interpretability analysis with the execution-guided strategy, our CESQL model demonstrates notable enhancements in both logical accuracy and execution accuracy on the WikiSQL dataset.

| Model | Dev Acc_lf | Dev Acc_ex | Test Acc_lf | Test Acc ex |
|---|---|---|---|---|
| IESQL+EG+AE | **87.9** | 92.6 | 87.8 | 92.5 |
| SDSQL+EG+AE | 86.7 | 92.5 | 87 | 92.7 |
| SeaD+EG_ACS | 87.6 | 92.9 | 87.5 | 93 |
| CESQL+EG+AE | 87.7 | **93.3** | **88** | **93.1** |

Table 2: Accuracy (%) of logic form Acc_lf and execution Acc_ex of our model CESQL and other competitors with EG decoding. Best results in bold. EG: execution-guided decoding. AE: rule-based aggregation enhancement. EG_ACS: the clause-sensitive EG strategy for S2S generation, with aggregation ignored during decoding.

Table 2 showcases the comparative results between our model and several others when employing rule-based aggregation enhancement on the WikiSQL dataset. Given that approximately 10% of the aggregation function labels in the WikiSQL dataset are incorrect, we adopted a data augmentation technique akin to Erosion for the training data. This technique involved removing some select column name information present in the query questions and randomly substituting some column names from different tables. By exploiting the dependency of aggregation functions on selected column names, we aimed to enhance the model's ability to adapt to predictions of incorrect aggregation functions. Our method was solely intended for the purpose of comparing the prediction outcomes of other models utilizing AE.

| Model | S_col | S_agg | W_col | W_op | W_val |
|---|---|---|---|---|---|
| SQLova+EG | 96.5 | 90.4 | 95.5 | 95.8 | 95.9 |
| X-SQL+EG | 97.2 | 91.1 | 97.2 | 97.5 | 97.9 |
| HydraNet+EG | 97.6 | 91.4 | 97.2 | 97.5 | 97.6 |
| IESQL+EG | 97.6 | 90.7 | 97.9 | 98.5 | 98.3 |
| SeaD+EG | **97.9** | **91.8** | 98.3 | 97.9 | 98.4 |
| CESQL+EG | 97.8 | 91.3 | **98.4** | **98.8** | **98.9** |

Table 3: Test accuracy (%) on WikiSQL test set for various clause components of SQL. The best results in bold. EG: execution-guided decoding. S_col: select column. S_agg: select aggregation. W_col: where condition column. W_op: where condition operator. W_val: where condition value.

Table 3 showcases the comparative results of our decomposition task against several other models on the WikiSQL dataset. The CESQL model, comprising interpretability analysis, execution-guided strategies, filtering adjustments, logical association corrections, and model fusion, markedly boosts the semantic parsing capabilities for condition columns, operators, and values within the WHERE clause of SQL statements, effectively realizing condition enhancement.

# 5 Conclusion

We have validated the feasibility of integrating interpretability analysis into our model using the WikiSQL dataset. Specifically for determining condition values in WHERE clauses, this approach not only mitigates the impact of manually annotated training data but also diminishes the reliance on data within the condition columns of database tables. This more neutral interpretation enhances the model's foundational capabilities and generalization in practical applications, making it better suited for users to perform tentative screenings without examining all data. We hope that this foundational exploration can offer fresh perspectives for research addressing complex queries and scenarios involving irregular data in databases. Furthermore, we consider the incorporation of autoencoder models in an auxiliary capacity, combined with autoregressive large models, as a promising approach to expedite the deployment of text-to-SQL applications.

Although capturing fine-grained key feature information via interpretability analysis has enhanced the accuracy and transparency of SQL query generation, it has also resulted in a notable prolongation of model processing time. Moving forward, we must strike a balance between accuracy and computational efficiency by investigating more viable strategies, refining model architecture, and minimizing unnecessary program invocations to boost overall performance. Additionally, we aim to explore the integration of LIME with SHAP to analyze feature interactions, with the aspiration of broadening its application scope.